\documentclass[sigconf, natbib]{acmart}
\usepackage{svg}
\usepackage{bbm}
\usepackage{adjustbox}
\usepackage{array}
\usepackage[table]{xcolor}
\usepackage{booktabs}%
\usepackage{caption} 
\usepackage{soul}
\usepackage[normalem]{ulem}
\usepackage{url}
\usepackage[linesnumbered,ruled,vlined]{algorithm2e}
\usepackage{setspace}
\SetKwInput{KwInput}{Input}
\SetKwInput{KwOutput}{Output}

\usepackage{amsmath}
\usepackage{amsfonts}
\usepackage{multirow}
\DeclareMathOperator*{\argmax}{arg\,max}
\DeclareMathOperator*{\argmin}{arg\,min}

\usepackage{pifont}
\usepackage{subfig}

\definecolor{red_pastel}{RGB}{210, 42, 42} 
\definecolor{blue_pastel}{RGB}{4, 54, 183} 
\definecolor{blue_mpl}{RGB}{31, 119, 180} 
\definecolor{orange_mpl}{RGB}{255, 165, 0} 

\begin{document}

\title{
End-to-end Early Classification of Time Series \\ in Non-Stationary Environments 
}

\author{Aurélien Renault}

\email{aurelien.renault@orange.com}
\affiliation{%
  \institution{Orange Research \& AgroParisTech}
  \city{Paris}
  \country{France}
}

\author{Alexis Bondu}
\email{alexis.bondu@orange.com}
\affiliation{%
  \institution{Orange Research}
  \city{Paris}
  \country{France}}

\author{Antoine Cornuéjols}
\email{antoine.cornuejols@agroparistech.fr}
\affiliation{%
  \institution{AgroParisTech UMR MIA-Paris}
  \city{Palaiseau}
  \country{France}}

\author{Vincent Lemaire}
\email{vincent.lemaire@orange.com}
\affiliation{%
  \institution{Orange Research}
  \city{Paris}
  \country{France}}

\begin{abstract}

Early Classification of Time Series (ECTS) requires making accurate decisions as early as possible in inherently online and evolving environments. Yet, most existing methods assume stationarity and rely on separable designs, where classification and triggering are optimized independently—an assumption that fundamentally limits their adaptability under drift.
In this work, we challenge this paradigm and study ECTS under non-stationary conditions. We provide the first systematic comparison between separable and end-to-end approaches across controlled drifting scenarios. Building on Reinforcement Learning, we introduce DQeND, a unified architecture that jointly learns representation, classification, and triggering decisions, while remaining directly comparable to state-of-the-art separable baselines.
Across a wide range of drifts, DQeND demonstrates strong robustness across various non-stationary scenarios, consistently outperforming separable baselines. An ablation study further highlights that jointly updating representation and decision modules is critical to these gains. Overall, our results indicate that end-to-end learning can offer improved adaptation capabilities for ECTS in dynamic environments, and motivate further investigation of alternatives to separable designs.

\end{abstract}



\maketitle

\section{Motivation}

\textit{Early Classification of Time Series} (ECTS) has emerged as a critical paradigm for applications where timely decision-making is as important as predictive accuracy \cite{gupta2020approaches, akasiadis2024framework, renaultearly}. By allowing a model to issue a classification before observing the full time series, ECTS systems inherently operate under a trade-off between \textit{earliness} and \textit{accuracy}. While substantial progress has been made in designing accurate early classifiers, most existing approaches are developed and evaluated under the implicit assumption of a stationary data-generating process. However, this assumption rarely holds in practical deployments, where the process generating successive time series can be subject to distribution shifts and concept drifts. Despite the growing maturity of the ECTS literature, the impact of non-stationarity remains largely unexplored. 

For instance, detecting toxic behavior on social media is subject to strong temporal constraints, as harmful content should be identified and filtered as early as possible. Moreover, the nature of such content can evolve with current events, e.g. during election periods, requiring early detection systems to continuously adapt to shifting definitions of what should be flagged. 

One specificity of ECTS is that it can be viewed as a dual decision process: (\textit{i}) the classification task, which aims to predict the class label based on partial observations, and (\textit{ii}) the trigger (or stopping) task, which determines the optimal time at which a prediction should be emitted. The vast majority of existing methods adopt a separable design, where these two components are implemented independently and optimized one after another \cite{gupta2020approaches, renaultearly}. Such a decomposition offers practical advantages, including flexibility, interpretability, and ease of optimization. However, it also implicitly assumes that classification and stopping decisions can be optimized independently. 

This assumption becomes particularly limiting when the underlying time series distribution evolves over time.
In such conditions, both the classification boundaries and the optimal stopping time may evolve in a coupled manner.
A separable framework, by design, may struggle to capture this joint adaptation, as each component is typically optimized in a sequential fashion, e.g., encoder $\rightarrow$ classifier $\rightarrow$ trigger, preventing upstream modules from benefiting from feedback provided by downstream decisions. In contrast, end-to-end architectures alleviate these unidirectional optimization dependencies by enabling the learning signal, namely the incurred cost, to propagate throughout the entire decision pipeline. Yet, this potential benefit of end-to-end learning for ECTS under distribution drift has yet to be investigated. Indeed, on one hand, existing work on end-to-end ECTS are limited in scope: they are often evaluated under stationary conditions and rarely benchmarked against strong separable baselines. On the other hand, existing work on non-stationarity remains in the separable scope and mainly focuses on adapting the triggering mechanism under changing decision costs, while assuming a fixed classification model \cite{renault2026early}. Consequently, a question  stands out: does jointly optimizing classification and stopping decisions improve robustness and adaptability when the underlying time series distribution itself evolves? And can we identify where the differences with separable methods, if any, manifest themselves? 

In this work, we provide a systematic investigation of separable and end-to-end ECTS architectures under controlled non-stationarities. Our contributions are threefold: 
\begin{itemize}
    \item We conduct an extensive empirical comparison between separable and end-to-end paradigms, explicitly analyzing the regimes in which joint optimization becomes beneficial.
    
    \item Building on recent work on Reinforcement Learning (RL)-based ECTS \cite{martinez2020adaptive, renault2025deep}, we propose an end-to-end framework that jointly optimizes classification and triggering decisions, while being directly comparable to state-of-the-art separable baselines.
    
    \item We design controlled experimental protocols involving covariate shifts and concept drifts, demonstrating that end-to-end optimization improves robustness and adaptation capabilities across a wide range of evolving environments.
\end{itemize}

Together, our results provide new insights into the design of adaptive ECTS systems and highlight the importance of jointly learning prediction and stopping decisions in dynamic environments. Experiments of this paper are fully reproducible and the source code can be found at \url{https://anonymous.4open.science/r/ects_concept_drift-8177}

The remainder of this paper is organized as follows. Section \ref{sec_background} provides background on ECTS. Section \ref{related_ects} reviews related ECTS works. Section \ref{methodology} describes the proposed RL-based end-to-end ECTS pipeline. Section \ref{sec_experimental_protocol} describes the experimental protocol, including the evaluation metrics, synthetic datasets, and the controlled cost drift scenarios. Finally, Section \ref{sec_results} presents and discusses the experimental results for several types of non-stationarities, and Section \ref{conclusion} concludes and discusses future work.

\section{Background}
\label{sec_background}

\subsection{Problem statement}

In the ECTS problem, an input time series of size $T$ is observed progressively. At time $t \leq T$, only a prefix ${\mathbf x}_t \, = \, \langle {x_1}, \ldots, {x_t} \rangle$ is available, where $\{x_i\}_{(1 \leq i \leq t)}$ denotes the measurements (possibly multivariate). The full series $\mathbf{x}_T$ belongs to an unknown class $y \in {\mathcal{Y}}$. The task is to output a prediction $\hat{y} \in {\mathcal{Y}}$ for the class of the incoming time series at some time $\hat{t} \in [1,T]$ before the deadline $T$.

ECTS explicitly addresses a trade-off between predictive accuracy and earliness. The prediction correctness is measured by a misclassification cost $\mathrm{C}_m(\hat{y}|y)$, and time pressure is modeled by a delay cost $\mathrm{C}_d(t)$, which is assumed to be positive and typically non-decreasing over time. Formally, we consider:  $\mathrm{C}_m : \mathcal{Y} \times \mathcal{Y} \rightarrow \mathbb{R}, \quad \mathrm{C}_d : \mathbb{R}^+ \rightarrow \mathbb{R}$.






Let $s \in \mathcal{S}$ be some ECTS function belonging to a class of functions $\mathcal{S}$, whose output at time $t$ when receiving $\mathbf{x}_t$ is:
\begin{equation}
s({\mathbf x}_t)  =  
\left\{
    \begin{array}{ll}
        \emptyset & \mbox{if extra measures are queried;}\\
        (\hat{y}_{t} ,\: t) & \mbox{if prediction is triggered at } t < T;\\    
    \end{array}
\right.
\label{eq:ects_model}
\end{equation}
Specifically, when the deadline is reached, i.e. $t=T$, the model is forced to return a prediction: $s(\textbf{x}_T) = (\hat{y}_T, T)$. \\
In practice, ECTS is an online optimization problem: at each time $t$, the function $s(\mathbf{x}_t)$ must decide whether to predict or wait for more data, based only on the prefix $\mathbf{x}_t$. 
The $s$ function induces a stopping time $\hat{t}$ on each incoming series that is the first time step for which $s$ returns a non-null prediction. It can  be defined as:
\begin{equation}
    \hat{t} = \inf\{ t \: | \: s(x_t) \neq \emptyset\}
    \label{eq:t_hat}
\end{equation}
Once triggered, the remaining observations $\{\textbf{x}_{t}\}_{(\hat{t} < t \leq T)}$ are not considered and prediction process is stopped.
The goal is to choose $\hat{t}$ such that the incurred, time-dependent, loss $\mathcal{L}(s(\textbf{x}_{\hat{t}}), y) = \mathrm{C}_m(\hat{y}_{\hat{t}}|y) + \mathrm{C}_d(\hat{t})$ is as small as possible.

From a machine learning perspective, the goal is to find a function $s \in {\mathcal{S}}$ that best optimizes the time-dependent loss function $\mathcal{L}$, minimizing the true risk over the data-generating process $\mathbb{P}{(\mathcal{X} \times \mathcal{Y})}$ \cite{renaultearly}:

\begin{equation}
        \argmin_{s \in \mathcal{S}} \mathbb{E}_{(\mathbf{x}_T, y) \sim \mathbb{P}_{(\mathcal{X} \times \mathcal{Y})}}\left[ {\mathcal{L}(s(\textbf{x}_{\hat{t}}), y)} \right]
\label{eq:cost1}
\end{equation} 

{In order to solve the ECTS problem, a training set is made available that is composed of $N$ labeled time series, denoted by $({\mathbf x}^i_T, y^i)_{i \in [0, N]} \in (\mathcal{X} \times \mathcal{Y})$, 
where each series ${\mathbf x}_T = \, \langle {x_1}, \ldots, {x_T} \rangle $ is complete and of the same size $T$, and associated with its label $y \in \mathcal{Y}$.}

\subsection{Online ECTS with drift}

The standard ECTS formulation typically assumes that the data-generating process observed at training time remains unchanged at deployment. This implicitly entails that both the input distribution and the underlying predictive relationship are stationary. In practice, this assumption is often violated: the \textit{actual} conditions encountered in production may differ from those seen during training and can evolve over time. Such variations may affect the input distribution (covariate shift), or the conditional relationship between inputs and labels (concept drift). 
Those may occur either within the duration of a single time series (i.e., along the time index $t \in [0, T]$) or across longer time scales, for instance due to seasonal or operational changes. 

To explicitly account for these different temporal dimensions, we distinguish between the time index of a given time series and the deployment timeline. Let $u \in \mathbb{N}^+$ index successive incoming labeled time series $\{ (\textbf{x}_T, y)^u \}_{(1\leq u \leq U)}$ observed during deployment, $(\textbf{x}_T, y)^u$ is generated by some distribution $\mathcal{D}_{u}$ over $\mathcal{X} \times \mathcal{Y}$. 

In the general case, the deployment process is characterized by several sources of uncertainty: \textit{(i)} the distribution $\mathcal{D}_{u}$ is \textit{unknown} to the learner at deployment; 
\textit{(ii)} it can be \textit{non-stationary}, i.e., drifting over deployment time $u$. 
%

At deployment, an ECTS system repeatedly processes incoming time series drawn from the evolving sequence of distributions ${\{\mathcal{D}_{u}\}}_{u \geq 1}$. 
The deployment performance of some ECTS function $s \in \mathcal{S}$ is then measured by: 
\begin{equation}
R_U(s) = \sum_{u=1}^U \mathbb{E}_{(\mathbf{x}_T,y)\sim \mathcal{D}_{u}}\big[\mathcal{L}(s(\textbf{x}_{\hat{t}}), y)\big],
\label{eq:deployment-loss-general}
\end{equation}
where the expectation is taken with respect to the current (and potentially drifting) data-generating process, as well as any additional sources of randomness. 
%
%

The remainder of the paper instantiates and compares different controlled forms of non-stationarity, encompassing data and concept drifts and evaluates ECTS methods' robustness across this wide range of dynamic scenarios.

\section{Related work} \label{related_ects}


The ECTS literature can be broadly divided into \emph{end-to-end} and \emph{separable} approaches. End-to-end methods jointly learn a classifier and a triggering mechanism. While flexible, they are classifier-specific and often difficult to compare consistently across studies. In contrast, \emph{separable} approaches decouple classification and decision triggering. This modular design dominates the literature and has been extensively evaluated \cite{gupta2020approaches,akasiadis2024framework,renaultearly}.

\subsection{Separable ECTS}

\paragraph{Classification module}

A common design choice in separable ECTS systems consists in discretizing the temporal axis and learning a collection of classifiers, each specialized for a given timestamp or subset of timestamps. This strategy enables the direct reuse of state-of-the-art Time Series Classification algorithms, while ensuring that predictions can be produced at different stages of the sequence. Among recent advances, the MiniROCKET pipeline \cite{dempster2021minirocket} has emerged as a particularly strong choice, combining high predictive performance with excellent scalability. Its effectiveness stems from the use of a large number of fixed random convolutional kernels, coupled with simple yet informative aggregation functions, yielding rich representations at a very low computational cost. As such, MiniROCKET-based classifiers are frequently adopted as backbone models in modern separable ECTS pipelines \cite{bilski2023calimera, renault2025deep, renault2026early}.

\paragraph{Trigger module}

On the decision side, a widely used baseline is the Proba Threshold strategy, which triggers a prediction as soon as the maximum posterior class probability exceeds a predefined threshold. Despite its simplicity, this approach has proven to be competitive across a variety of datasets and cost configurations, and is therefore often used as a reference method \cite{renaultearly}.

Beyond such myopic strategies, most high-performing trigger models fall into the class of \textit{non-myopic} approaches, which explicitly account for the potential evolution of future observations when making decisions. The Economy family of methods \cite{achenchabe2021early, dachraoui2015early} follows this principle by triggering a decision when the expected cost at the current time step is minimal among all anticipated future costs. Similarly, Calimera \cite{bilski2023calimera} adopts an anticipation-based framework, leveraging regression models learned via backward induction to estimate future costs and guide the triggering decision. 

Recently, RL has been introduced as a flexible alternative for modeling the triggering process. In particular, Alert \cite{renault2025deep} learns a triggering policy, based on a state space composed by handcrafted features, obtained by probing the ensemble of calibrated MiniROCKET estimators. In addition to strong empirical performance, this approach is inherently well-suited to online adaptation and can naturally incorporate feedback from a deployment phase. For those reasons, the present work primarily builds upon this trigger model and investigates its integration within an end-to-end architecture, where the classifier is directly informed by the triggering signal.


\subsection{End-to-end ECTS}


Historically, the earliest works that can be considered end-to-end ECTS consist of a classifier-specific stopping criterion, directly integrated within a base Time Series Classification algorithm. A seminal example is the work of \cite{xing2009early}, which determines a unique minimum prediction length such that the predictions of a 1-nearest neighbor classifier remain the same as those obtained using a complete time series. In a similar vein, the EDSC algorithm \cite{xing2011extracting} builds upon shapelet-based classification by associating each shapelet with a temporal penalty, thereby implicitly favoring patterns that occur earlier in the time series. 

More recent advances in deep learning have enabled tighter integration between representation learning, classification, and decision timing. In particular, ELECTS \cite{russwurm2023end} proposes an LSTM-based architecture, augmented by a stopping head that determines the probability of triggering. It is trained with a dedicated loss function that explicitly balances classification accuracy and earliness. 

In parallel, the growing popularity of RL has led to a number of approaches that formulate ECTS as a sequential decision-making problem.
Earlier work by \cite{martinez2018deep, martinez2020adaptive} introduced the use of Deep Q-Networks (DQN) for ECTS, modeling the triggering task as a value-based decision process. EARLIEST \cite{hartvigsen2019adaptive, hartvigsen2020recurrent} relies on a policy gradient method to learn a stopping policy jointly with a LSTM-based encoder and classifier. 
These RL-based methods offer a natural framework to capture the trade-off between earliness and accuracy, while enabling flexible adaptation to complex and potentially non-stationary environments.



\section{DQeND: End-to-end RL for ECTS} \label{methodology}


In this section, the design choices for the proposed end-to-end ECTS pipeline, called DQeND, are described. 
In contrast to traditional separable approaches, where the different components are optimized independently, our architecture integrates them into a single decision process that can be trained and adapted holistically. The pipeline consists of two main modules: (\textit{i}) an \textit{encoder} that transforms raw time series prefixes into informative latent representations, and (\textit{ii}) a RL–based \textit{decision} module that simultaneously determines when to stop and which class to predict.

\begin{figure*}
    \centering
    \includegraphics[width=0.85\linewidth]{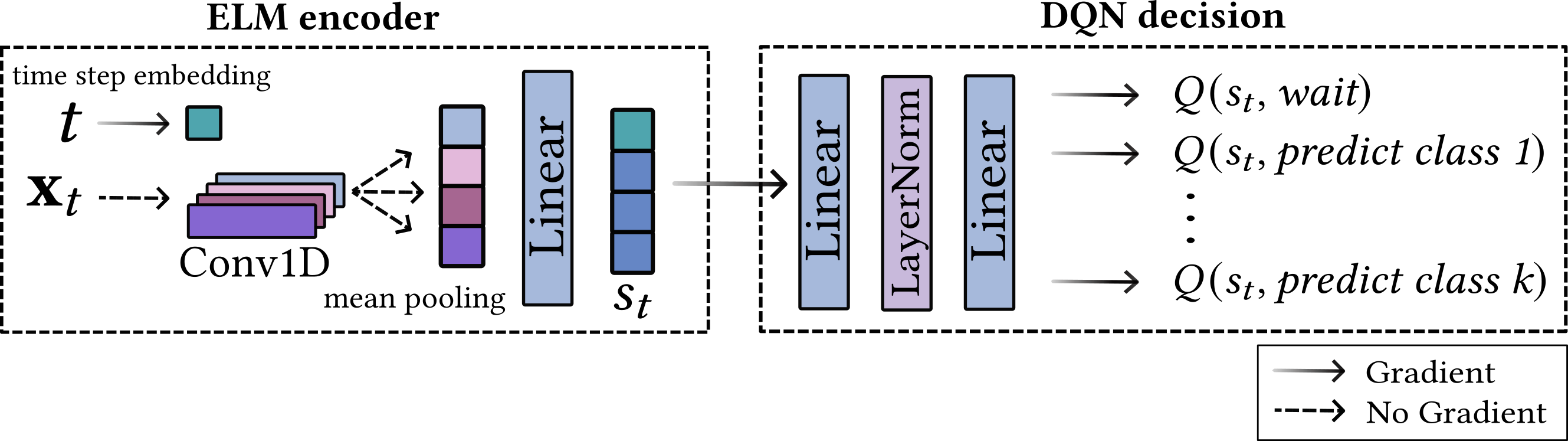}
    \caption{DQeND pipeline}
    \label{fig:pipe}
\end{figure*}

To enable fair comparisons with state-of-the-art separable methods and isolate the impact of end-to-end learning from architectural differences, we build our pipeline upon both ROCKET-type and Alert frameworks \cite{dempster2019rocket, dempster2021minirocket, renault2025deep}. The global DQeND pipeline is illustrated in Figure \ref{fig:pipe}.

\subsection{Architecture}

\subsubsection{ELM Encoder} \label{encoder}

The design of the encoder module is guided by the objective of combining strong representational capacity with computational efficiency in an online setting. To this end, we draw inspiration from the principles underlying the widely adopted, ROCKET family of methods \cite{dempster2019rocket, dempster2021minirocket}, whose features are computed via a large number of random convolutions. More broadly, the idea of generating a large hidden representation of the input data using random weights originates from the Extreme Learning Machine (ELM) literature \cite{huang2006extreme, wang2022review}. 

At a high level, ELM-based approaches are single-layer neural networks. They rely on projecting input data into a high-dimensional feature space using a large set of randomly initialized and fixed parameters, and subsequently learning a lightweight mapping on top of this representation. Formally, the output of some ELM network with $L$ hidden nodes can be defined as:

\begin{equation}
    f_L(\textbf{x}) = \sum_{i=1}^L \beta_i h_i(\textbf{x})
\end{equation}

where $\beta  = [\beta_1, \cdots, \beta_L]$ is the learned weight matrix, and $\textbf{h}(\textbf{x}) = [h_1(\textbf{x}), \cdots, h_L(\textbf{x})]$ the fixed random features for the $i$-th hidden node. 

This paradigm has been shown to yield competitive performance across a wide range of time series classification tasks \cite{liu2016extreme, park2017online}, while maintaining a favorable computational profile due to the limited number of trainable parameters. Such properties are particularly desirable in online learning scenarios, where frequent updates must be performed under efficiency constraints. 

Combining the ELM principles with the ROCKET-type of time series transformations, the proposed encoder implements a randomized feature extractor based on a single one-dimensional convolutional layer with frozen weights. Multiple kernel sizes $k$ are employed in parallel in order to capture temporal patterns at different scales, thereby enriching the expressiveness of the resulting representation. The convolutional outputs are then aggregated through a mean pooling operation, producing a fixed-size embedding regardless of the input length. In particular, feature $h_i$ can be obtained through the following operations: \\
\begin{equation}
    h_i(\textbf{x}_t) = \text{AvgPool(}\text{Conv1D}_k(\textbf{x}_t))
\end{equation}

The convolutional filters are initialized by sampling their weights from a standard normal distribution and remain unchanged throughout training and deployment, ensuring a stable and inexpensive feature extraction process. To complement this representation, we additionally learn an explicit embedding of the time step, which is concatenated to the ELM-based features. This allows the downstream decision modules to incorporate temporal position information alongside the extracted signal patterns, a key aspect for early classification tasks where the decision process is inherently time-dependent.


\subsubsection{DQN Decision}

The decision-making module is designed to remain directly comparable to strong separable baselines while enabling full end-to-end optimization. To this end, we adopt the same underlying architecture as the Alert framework \cite{renault2025deep}, namely a Deep Q-Network (DQN)-based decision model \cite{dayan1992q, sutton2018reinforcement}. This choice ensures that any observed differences in performance can be attributed to the representation and training paradigm, rather than to architectural discrepancies in the decision process itself. 

In contrast to separable approaches, Alert in particular, that typically rely on handcrafted features or classifier-derived confidence scores, the proposed module operates directly on the learned representation produced by the encoder. More precisely, the input state $s_t$ is given by the embedding output at time $t$, allowing the decision module to leverage rich, task-adaptive features without additional engineering. 

The action space is extended so as to jointly model the stopping decision and the class prediction. Specifically, we define $\mathcal{A} = \{ \text{``\textit{wait}''}, \text{``\textit{predict class 1}''}, \cdots, \text{``\textit{predict class }$K$''} \} $, there\-by unifying the triggering and classification tasks within a single decision process. At each time step, the selected action follows the standard greedy policy induced by the Q-function:  

\begin{equation}
    a_t = \arg\max_{a \in \mathcal{A}} Q(s_t, a)
\end{equation}

The prediction mechanism is derived directly from the chosen action. If the model selects a prediction action, the corresponding class is immediately output. 
A specific handling is introduced for the terminal time step $T$: if the selected action remains ``\textit{wait}'', the model is forced to emit a prediction, corresponding to the class associated with the highest Q-value among the non-waiting actions. Formally, the predicted label is defined as:
\begin{align}
    \hat{y}_t(a_t) &= \left\{
    \begin{array}{ll}
         \emptyset & \mbox{if } a_t=\text{``\textit{wait}''} \mbox{ and } t<T \\ 
         \displaystyle \argmax_{\footnotesize a\in\mathcal{A} \setminus \{\text{``\textit{wait}''} \}} Q(s_t, a) & \mbox{if }  a_t=\text{``\textit{wait}''} \mbox{ and } t=T \\
         a_t & \mbox{if }  a_t \neq \text{``\textit{wait}''} \\  
    \end{array}
    \right.
\end{align}
%


\subsection{Training (offline)}

For stability reasons, learning is performed offline in a batch setting, where the model is trained on a fixed collection of fully labeled time series without requiring active exploration during training. In particular, the overall training procedure is kept close to that of Alert \cite{renault2025deep}.
 
%

The RL objective is defined through a reward function that captures the trade-off between earliness and accuracy. At each time step, the reward is expressed as the temporal variation of the delay cost when the agent decides to wait, and additionally incorporates the misclassification cost when a prediction is made. Formally, the reward is defined as: 
\begin{align}
    r(t) &= \left\{
    \begin{array}{ll}
         -\Delta \mathrm{C}_d(t) & \mbox{if } a_t=\text{``\textit{wait}''} \\  
         -C_m(\hat{y}_t|y) - \Delta \mathrm{C}_d(t) & \mbox{if } a_t \neq \text{``\textit{wait}''}
    \end{array}
    \right.
    \label{eq:r1}
\end{align}
where $-\Delta \mathrm{C}_d(t) = \mathrm{C}_d(t) - \mathrm{C}_d(t-1)$, with $\mathrm{C}_d(0) = 0$. 
%
%

The gradient of the DQN objective, corresponding to the Bellman error, is propagated through the entire DQeND pipeline, updating all parameters associated with the shaded arrows in Figure \ref{fig:pipe}.
Finally, the model selection procedure is kept identical to that of \cite{renault2025deep}.




\subsection{Deployment (online)}

The proposed framework naturally extends to an online adaptation setting, where both the encoder and the decision modules are jointly updated based on the feedback induced by the decisions. In contrast to separable approaches, this joint update enables the representation itself to evolve in response to non-stationary conditions, allowing for a tighter coupling between feature extraction and decision-making. 

At deployment time, each observed interaction is stored in a replay buffer. The pipeline is then updated by sampling mini-batches of transitions from this buffer, following standard DQN optimization procedures. 

When moving from offline training to online adaptation, exploration is explicitly reintroduced in order to account for potential distribution shifts. Concretely, we adopt an $\epsilon$-greedy strategy, where with probability $\epsilon$ a random action is selected, and with probability $1 - \epsilon$ the greedy action, i.e. the one with maximum $Q$-value, is chosen. In the exploration phase, actions are sampled uniformly from the action space, ensuring equal probability across all possible decisions (e.g., $1/3$ for each action in the binary classification case). In all experiments, the exploration rate is fixed to $\epsilon = 5\%$. Details on all the hyperparameters' values can be found in Table \ref{tab:hyperparams}. 
%


\begin{table}[!htb]
    \centering
    \caption{DQeND hyperparameters}
    \begin{tabular}{c|cc}
        \hline
         \multirow{4}{*}{\textbf{Encoder}} & kernel sizes & $\{3,6,9,12\}$ \\
         & nb. of kernels & $\{128,64,32,16\}$ \\
         & embedding dim. & 64 \\
         & time embedding dim. & 2 \\
         \hline
         \multirow{3}{*}{\textbf{Decision}} & nb. hidden layers & 1 \\
         & hidden dim. & 64 \\
         & activation function & tanh \\
         \hline
         \multirow{3}{*}{\textbf{Training}} & learning rate & 1e-3 \\
         & nb. epochs & 100 \\
         & batch size & 256 \\
         \hline
         \multirow{3}{*}{\textbf{Deployment}} & learning rate & 1e-4 \\
         & exploration rate $\epsilon$ & 0.05 \\
         & batch size & 64 \\
         \hline
    \end{tabular}
    \label{tab:hyperparams}
\end{table}

\section{Experimental Protocol}
\label{sec_experimental_protocol}

This section describes the experimental protocol used to evaluate the proposed online ECTS methods under non-stationarities. We detail the evaluation metrics, the data generation process, the drifts scenarios, and the evaluated methods. All these elements are fixed prior to the analysis of the results presented in Section~\ref{sec_results}. 

\subsection{Evaluation Metrics}

All online methods are evaluated using an \emph{eval-then-update} (prequential) protocol~\cite{10.1145/1557019.1557060}. At each deployment step $u$, a time series and its associated deployment-time costs are first used to evaluate the current model. The observed loss is then used to update the models according to its update strategy.

{
To study the impact of different trade-offs between misclassification and delay costs, we use the standard weighted formulation of the average cost, along with the cost functions definitions used by \cite{renaultearly, achenchabe2021early, bilski2023calimera}:
\begin{align*}
    \textit{AvgCost}_{\alpha} &= \frac{1}{N} \sum_{i=1}^{N} \mathrm{C}_m(\hat{y}_i \mid y_i) + \mathrm{C}_d(\hat{t}_i) \\
    &= \frac{1}{N} \sum_{i=1}^{N} \alpha\, \mathbbm{1}(\hat{y} \neq y) + (1-\alpha)\, \frac{t}{T}
    \label{eq_avgcost_weight}
\end{align*}
where $\alpha \in [0,1]$ controls the relative importance of accuracy and earliness.}
{
Throughout this work, as in \cite{mori2017reliable, lv2023second}, we fix $\alpha = 0.8$.
}

The evaluation relies on a set of three metrics derived from the standard \textit{AvgCost}, in order to capture complementary aspects of performance under non-stationary conditions: 

\begin{itemize}
    \item First, we report the \textbf{\textit{cumulative cost}}, defined as the total sum of incurred costs over all deployment examples. 
    \item Second, we consider the \textbf{\textit{hold-out AvgCost drift}}, computed on a disjoint evaluation set that is subject to the same drift conditions as the current deployment step. Unlike the raw cumulative deployment cost, which may exhibit high variance due to transient effects or small batches, this metric provides a more stable estimate of performance under the current time series distribution.
    \item Finally, we report the \textbf{\textit{hold-out AvgCost original}}, evaluated on a separate hold-out set drawn from the original (training-time) distribution. This complementary metric assesses the extent to which a method preserves performance on the base concept, thereby offering insight into its ability to balance adaptation and retention; often referred to as the stability–plasticity trade-off \cite{kim2023stability, serra2025incremental}.
\end{itemize}




\subsection{Data Generation}



Studying non-stationarity in ECTS requires benchmarks where the difficulty of the task and the nature of the drift can be precisely controlled. In particular, the ability to independently manipulate factors such as noise intensity, temporal location of discriminative patterns, or class structure is essential. The MNIST-1D \cite{greydanus2020scaling} dataset naturally fulfills these requirements through its synthetic generation process, making it particularly well-suited for the controlled drift scenarios considered in this work.
The MNIST-1D dataset is a ten-class classification problem, which has been binarized in this work, based on univariate time series of fixed length $T=100$.  Each generated time series contains a class-specific pattern representing the shape of a digit, embedded at a random position.
Training phase is conducted using 1,000 examples (700 for training and the 300 left to compute the hold-out \textit{AvgCost} original); at each deployment step $u$, 1{,}500 new (drifted) examples are generated, among which 500 are used for updates and 1{,}000 to compute the hold-out \textit{AvgCost} drift. 



\subsection{Drift Scenarios}

To evaluate the robustness of online ECTS methods to non-stationarities, we consider controlled variations of input data and concept. Non-stationary patterns occur across deployment steps $u$, while the time index $t \in [0,T]$ remains associated with the progression within an individual time series. \\

{We first distinguish two temporal patterns of variation:
\begin{itemize}
    \item \textbf{Incremental (\textit{I}) drifts} occur when the drifted variable(s) evolve smoothly during deployment.
    \item \textbf{Abrupt (\textit{A}) drifts} occur when the drifted variable(s) suddenly switch from the training regime to a new deployment regime.
\end{itemize}
}

We consider three variables over which to intervene in order to generate drifts: 
\begin{itemize}
    \item \textbf{Template Location (\textit{loc})} defines the location of the true class pattern within a time series.
    \item \textbf{Gaussian Noise (\textit{noise})} defines the scale of the Gaussian noise to be applied over a time series.
    \item \textbf{Class shuffling (\textit{class})} defines the set of sub-classes characterizing the two binary (meta-)classes. 
\end{itemize}

Combining these two dimensions yields four experimental scenarios:
\begin{enumerate}
    \item \textbf{\textit{I\_loc}}: the class patterns are first located at the very beginning of the time series; it progressively shifts so that it reaches the end of the series at the end of deployment.

    \begin{figure}[!htb]
        \centering
        \includegraphics[width=0.9\linewidth]{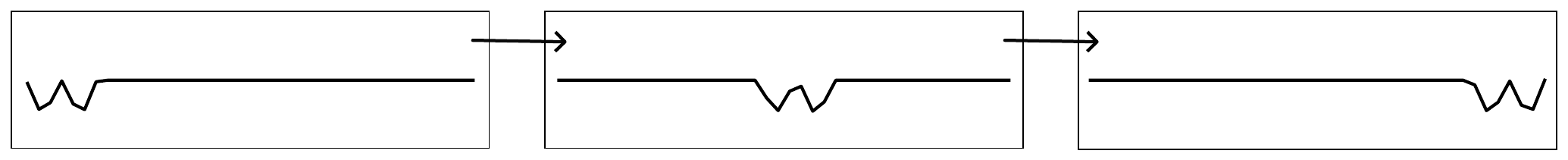}
        \caption{The \textbf{\textit{I\_loc}} scenario.}
        \label{fig:placeholder}
    \end{figure}
    
    \item \textbf{\textit{I\_noise}}: the scale of the \textit{iid} Gaussian noise becomes progressively greater, following a log spaced schedule from $0.01$ to $1$.

    \begin{figure}[!htb]
        \centering
        \includegraphics[width=0.9\linewidth]{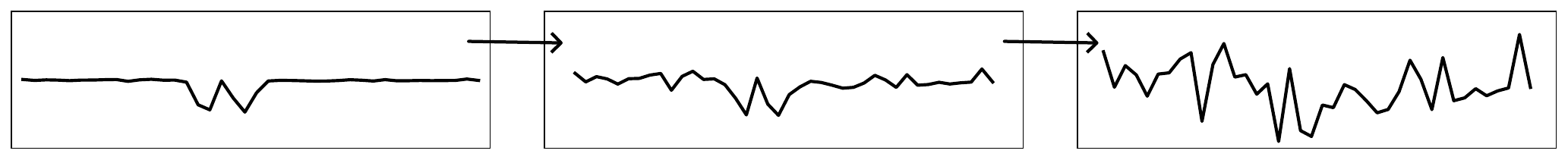}
        \caption{The \textbf{\textit{I\_noise}} scenario.}
        \label{fig:placeholder}
    \end{figure}
    
    \item \textbf{\textit{A\_loc}}: the class patterns are first located within the first quarter of the series; after five deployment steps, they are shifted to the last quarter.
    \vspace{1mm}
    \begin{figure}[!htb]
        \centering
        \includegraphics[width=0.66\linewidth]{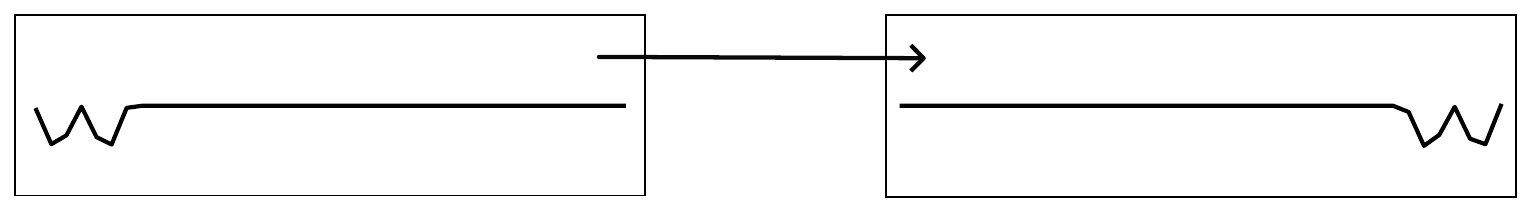}
        \caption{The \textbf{\textit{A\_loc}} scenario.}
        \label{fig:placeholder}
    \end{figure}
    
    \item \textbf{\textit{A\_class}}: the set of sub-classes defining the two main classes is completely resampled, with a $5\%$ probability of occurring at each deployment step. 


    \begin{figure}[!htb]
        \centering
        \includegraphics[width=0.95\linewidth]{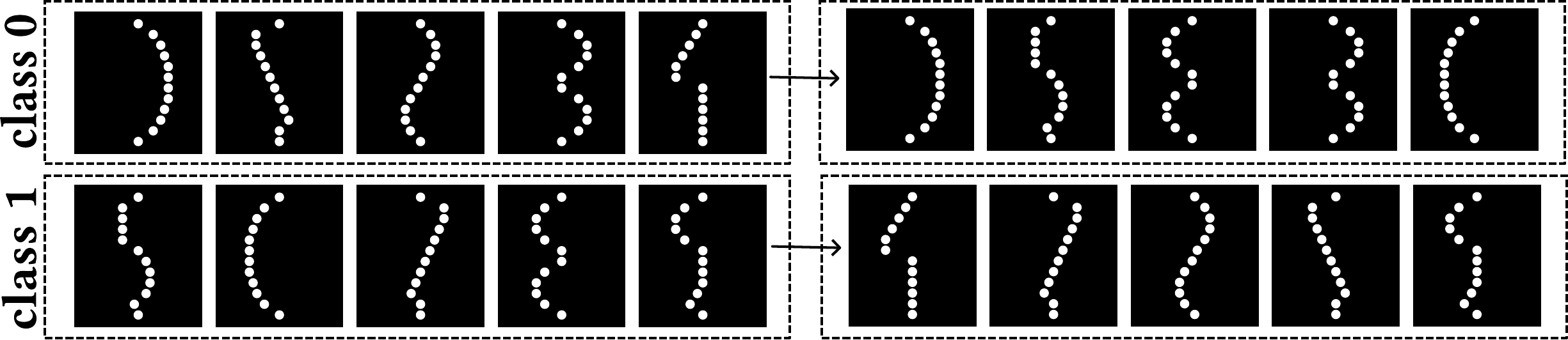}
        \caption{The \textbf{\textit{A\_class}} scenario.}
        \label{fig:placeholder}
    \end{figure}
    
\end{enumerate}

\subsection{Evaluated Methods}

\paragraph{Separable baselines} 


As highlighted in Section \ref{methodology}, Alert constitutes the comparison reference for the proposed end-to-end method. Thus, all the separable models described in this section are based on: (\textit{i}) a \textit{classifier} that shares the same architecture as the ELM encoder described in Section \ref{encoder}, followed by a linear classification layer and (\textit{ii}) a \textit{trigger} which is Alert, as described in the original paper \cite{renault2025deep}, including the handcrafted input space. In what follows, several ways of adapting separable baselines to online environments, depending on the addressed drift type, are presented.

\begin{itemize}
    \item \textbf{Common baselines}: \\
    (\textit{i}) \textit{bronze}: both the classifier and the trigger model remain frozen during deployment. \\
    (\textit{ii}) \textit{silver}: both the classifier and the trigger model are being trained on the concept on which they will be evaluated. Act as an upper bound, not realistic in practice.

    \item \textbf{Incremental drifts}: \\
    (\textit{i}) \textit{online/freeze}: the classifier is updated online, after each deployment step, the trigger model remains frozen. \\
    (\textit{ii}) \textit{online/online}: both the classifier and the trigger model are updated online, after each deployment step. The data batch is split in two so that each module is updated based on one half of the batch.

    \item \textbf{Abrupt drifts}: \\
    (\textit{i}) \textit{retrain/freeze}: the classifier is retrained from scratch, at a given frequency, the trigger model remains frozen. \\
    (\textit{ii}) \textit{retrain/retrain}: both the classifier and the trigger model are retrained from scratch, at a given frequency. The data batch is split in two so that each module is trained on the basis of one half of the batch.
\end{itemize}



\paragraph{End-to-end}

The following end-to-end methods have been adapted so that they may be updated online: ELECTS \cite{russwurm2023end} and EARLIEST \cite{hartvigsen2019adaptive}.

\subsection{Implementation details}

The deployment phase consists of $U=100$ steps of 500 examples each. For the \textit{retrain} baselines, the retraining frequency has been set to 20. 
For both ELECTS and EARLIEST algorithms, training phase is conducted over 500 epochs, the deployment batch size has been set to 64 and the learning rate to 5e-3. All experiments have been run on CPU, with up to five threads to perform cross-validation, when needed. Results that follow consist of the mean taken over five different random seeds. For all other details, we make the code available at: \url{https://anonymous.4open.science/r/ects_concept_drift-8177}

\section{Results}
\label{sec_results}

In this section, we analyze the ECTS online performances under the two types of drifts presented in Section \ref{sec_experimental_protocol}.


\subsection{Incremental drifts}

\paragraph{Incremental Location (\textbf{I\_\textit{loc}})}

This first experiment investigates a gradual shift of discriminative patterns toward later positions in the deployment time series.

\begin{figure}[!htb]
    \centering
     \subfloat[\textbf{\textit{I\_loc}}]{\includegraphics[width=0.5\linewidth]{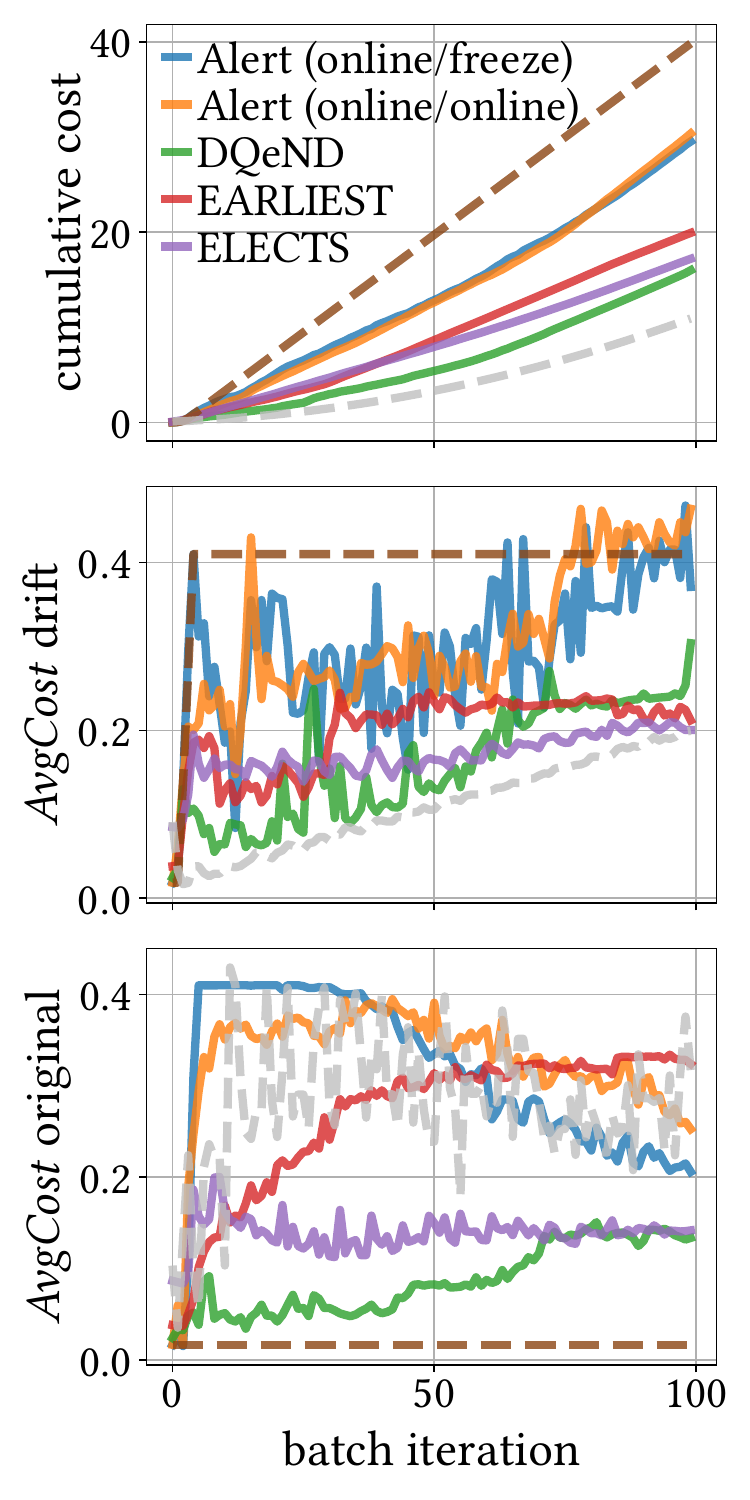}\label{fig:iloc}}
     \subfloat[\textbf{\textit{I\_noise}}]{\includegraphics[width=0.5\linewidth]{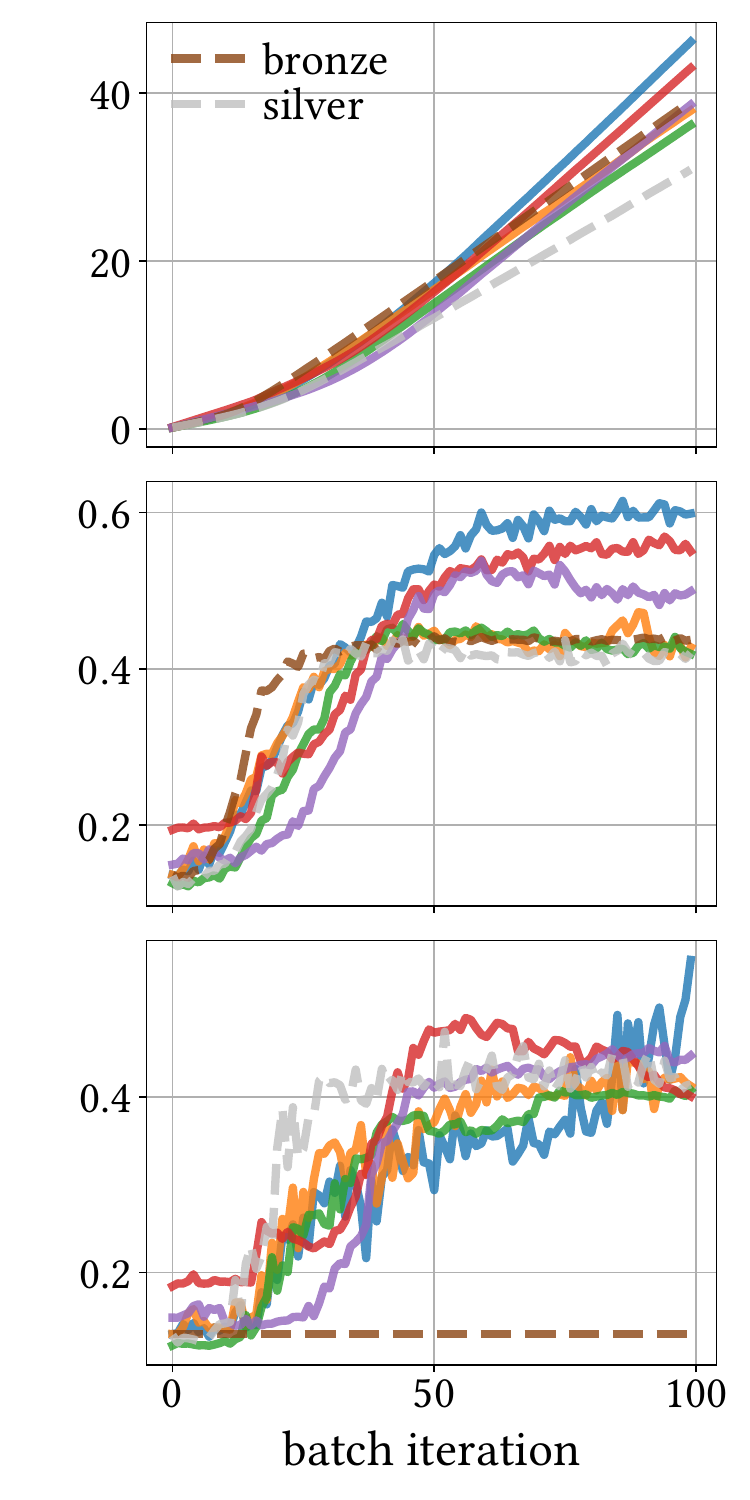}\label{fig:inoise}}
    \caption{Incremental drifts, deployment's metrics, averaged over five seeds. From top to bottom: the raw cumulative cost; the hold-out \textit{AvgCost}, obtained on unseen drifted data; the hold-out \textit{AvgCost} original, obtained using the fixed test set from training phase. UPPERCASE approaches refers to end-to-end methods while lowercase indicates separable ones.}
    \label{fig:incremental}
\end{figure}

\begin{figure}[!htb]
    \centering
      \subfloat{\includegraphics[width=1\linewidth]{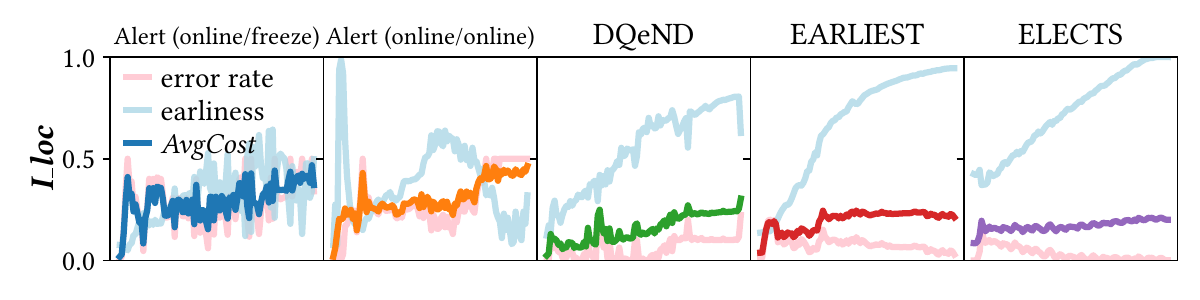}} \\ 
     \vspace{-5.8mm}
     \subfloat{\includegraphics[width=1\linewidth]{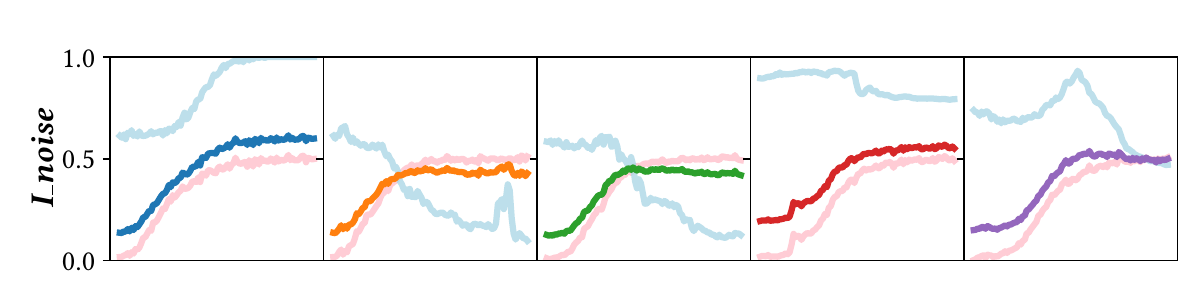}}
    \caption{Hold-out \textit{AvgCost} drift: detailed performances. The figures display decomposition of the hold-out \textit{AvgCost} drift into both earliness, i.e. the average proportion of the time series used, and error rate, during the deployment phase.}
    \label{fig:incr_details}
\end{figure}

As shown in Figure \ref{fig:iloc}, end-to-end approaches clearly outperform separable baselines, consistently achieving the lowest cumulative costs (top panel). In particular, DQeND ranks first overall and, together with ELECTS, best preserves performance on the original concept (bottom panel), demonstrating a strong balance between adaptation and retention. In addition, the qualitative analysis in Figure \ref{fig:incr_details} (top panel) shows that end-to-end models effectively adapt their triggering policies by progressively delaying predictions, illustrated by the increasing light-blue earliness curve of each box, as the discriminative patterns shift later in the series. 

In contrast, both Alert-based separable methods degrade significantly as informative patterns move later in the sequences (middle panel, right), highlighting the limitations of decoupled classification and triggering under temporal drift. Freezing the Alert trigger further leads to unstable behavior throughout deployment, whereas updating both classifier and trigger slightly improves stability but still fails to improve global performance.


\paragraph{Incremental Noise (\textbf{I\_\textit{noise}})}

This second experiment evaluates robustness under a gradual increase of additive Gaussian noise, progressively degrading the signal-to-noise ratio and the reliability of discriminative patterns.

As displayed in Figure \ref{fig:inoise}, end-to-end approaches again achieve the best cumulative costs, with DQeND ranking first, closely followed by ELECTS. Despite similar overall performance, analysing Figure \ref{fig:incr_details}, bottom panel, indicates that the two methods exhibit distinct adaptation behaviors. ELECTS follows a conservative strategy, initially delaying predictions to preserve accuracy while the signal remains informative, before progressively shifting toward earlier decisions as noise becomes dominant. In contrast, DQeND rapidly converges toward an early-prediction regime once accuracy approaches random performance levels (explaining the plateau shape in Figure \ref{fig:inoise}, middle panel), effectively minimizing delay costs when waiting no longer provides useful information. This difference in the adaptation strategy is illustrated by a vertical shift of the earliness curve of both algorithms. 

Regarding separable approaches, the same trends as in the incremental location drift experiment can be observed. 
In particular, the frozen Alert trigger again produces unstable behavior, whereas updating both classifier and trigger significantly improves robustness. The \textit{online}/\textit{online} variant achieves performance close to DQeND under low to moderate noise levels (Figure \ref{fig:inoise}, middle panel, left), although the end-to-end model remains slightly more robust, suggesting that unified architectures better exploit weak residual signal under mild perturbations. \\


\subsection{Abrupt drifts}

\paragraph{Abrupt Location (\textbf{\textit{A\_loc}})}

This experiment evaluates an abrupt drift scenario in which the discriminative pattern is suddenly shifted from the beginning to the end of the time series. 

\begin{figure}[!htb]
    \centering
     \subfloat[\textbf{\textit{A\_loc}}]{\includegraphics[width=0.5\linewidth]{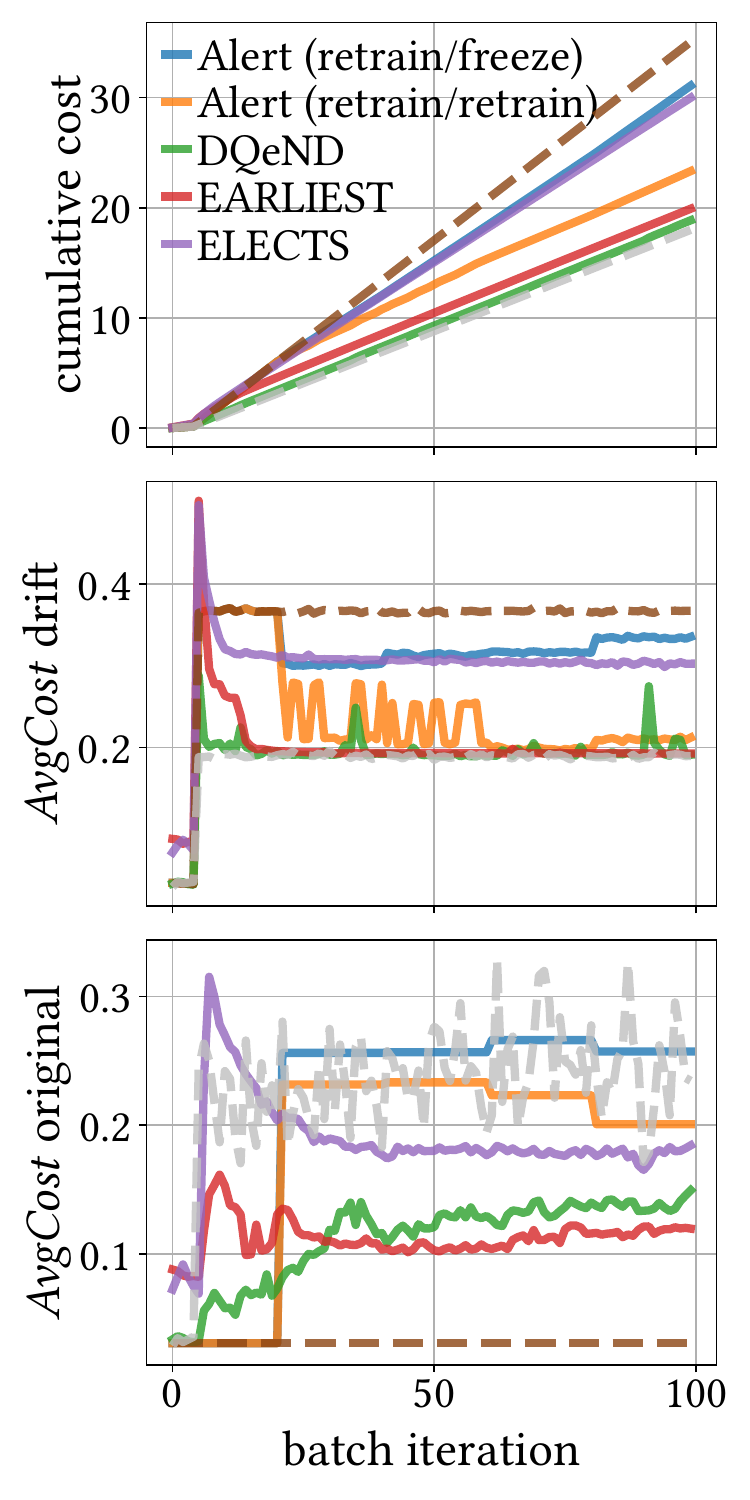}\label{fig:aloc}}
     \subfloat[\textbf{\textit{A\_class}}]{\includegraphics[width=0.5\linewidth]{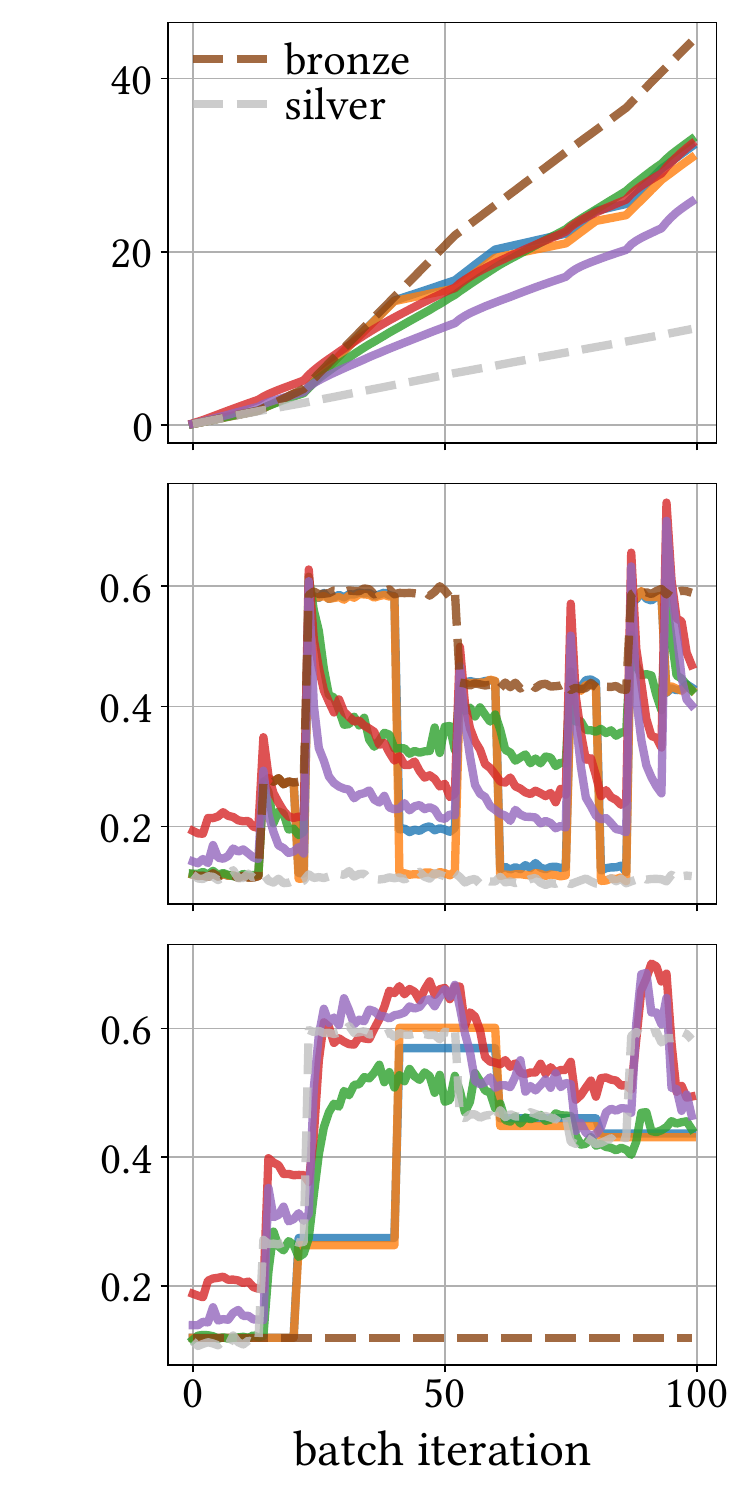}\label{fig:aclass}}
    \caption{Abrupt drifts, deployment's metrics, averaged over five seeds. From top to bottom: the raw cumulative cost; the hold-out \textit{AvgCost}, obtained on unseen drifted data; the hold-out \textit{AvgCost} original, obtained using the fixed test set from training phase. UPPERCASE approaches refers to end-to-end methods while lowercase indicates separable ones.}
    \label{fig:incremental}
\end{figure}

As shown in Figure \ref{fig:aloc}, top-middle panel, the RL-based approaches adapt most effectively to this sudden change. Both DQeND and EARLIEST rapidly shift from early to late predictions, successfully adjusting their triggering policies to the new pattern location. RL-based methods also best preserve performance on the original concept (Figure \ref{fig:aloc}, bottom), illustrating their ability to balance adaptation and retention. In contrast, ELECTS 
struggles to substantially delay its decisions at deployment, with trigger points rarely extending beyond half of the sequence. This limited shift highlights a lack of plasticity, as the model remains biased toward its initial regime and fails to fully adapt to the new optimal decision timing.

For the Alert-based separable baselines, the importance of online adaptation becomes even more evident. Freezing the trigger model can lead to catastrophic failures, as the decision policy becomes completely misaligned with the shifted distribution. Updating the trigger substantially improves stability and eventually recovers appropriate triggering behavior, although adaptation remains slower than with end-to-end approaches.


\paragraph{Sub-classes shuffling (\textbf{\textit{A\_class}})}

This final experiment considers a highly non-stationary setting where the composition of meta-classes is randomly reshuffled at irregular intervals, directly altering the underlying class structure rather than the time series distribution. 

\begin{figure}[!htb]
    \centering
     \subfloat{\includegraphics[width=1\linewidth]{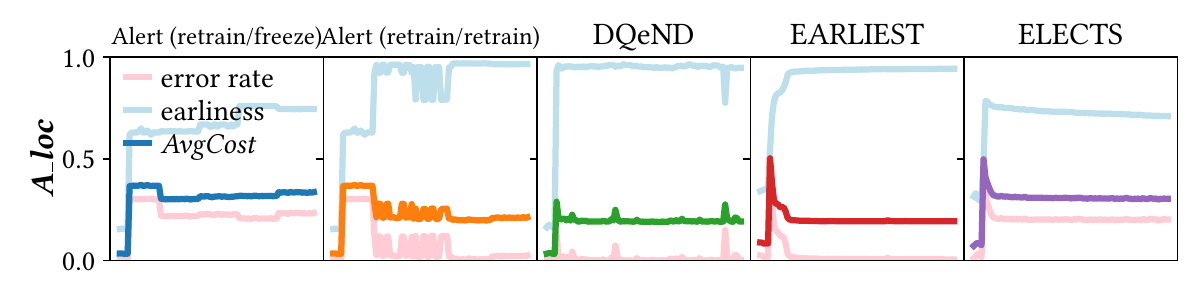}} \\ 
     \vspace{-5.8mm}
     \subfloat{\includegraphics[width=1\linewidth]{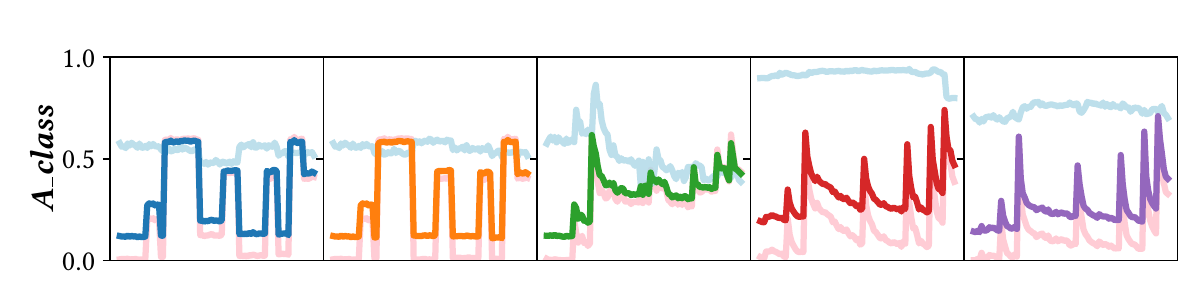}}
    \caption{Hold-out \textit{AvgCost} drift: detailed performances. The figures display decomposition of the hold-out \textit{AvgCost} drift into both earliness, i.e. the average proportion of the time series used, and error rate, during the deployment phase.}
    \label{fig:abrupt_details}
\end{figure}

In Figure \ref{fig:aclass}, ELECTS achieves the best overall performance, consistently recovering after each drift event. The RL-based approaches, DQeND and EARLIEST, closely follow and remain competitive. Noticeably, RL-based approaches suffer from a stronger constraint: unlike ELECTS and separable baselines, they only observe the data up to the estimated trigger time rather than the full time series. This makes adaptation to abrupt-type drifts substantially more challenging, a strong exploration rate being thus critical to efficiently recover. A more detailed analysis via Figure \ref{fig:abrupt_details}, bottom panel, nevertheless reveals a limitation of DQeND in this regime. Under repeated abrupt drifts, its triggering policy progressively collapses toward systematically early predictions, represented by the light-blue earliness curve, which keeps decreasing. This highlights the instability of the fully RL-based pipeline, which likely results from the accumulation of policy updates across successive distribution shifts. In contrast, ELECTS and EARLIEST display more stable earliness profiles (almost flat earliness curve) throughout deployment, reflecting a more conservative adaptation strategy. 

Within this setting, separable baselines achieve performance comparable to the RL-based approaches, although being retrained at fixed, predetermined intervals. Interestingly, updating the trigger model brings only limited improvements over keeping it frozen.


\subsection{Ablation studies}

To further assess the benefits of the proposed end-to-end design, we conduct an ablation study in which the encoder and decision modules of DQeND are alternatively frozen during adaptation. This allows isolating the respective contributions of representation learning and decision-making in the presented non-stationary settings. 

The results, reported in Table \ref{tab:ablation}, show that jointly updating both modules consistently yields the best performance across most scenarios. In contrast, partially frozen variants lead to noticeable degradations, confirming the importance of coordinated adaptation. Among the two, freezing the decision module results in the most significant performance drop.
This observation is further supported by the fact that the decision-making module accounts for most of the learnable parameters in the proposed architecture. 

Interestingly, freezing the encoder has a more limited impact in the abrupt drift settings considered in our experiments. This suggests that a well-trained representation can remain sufficiently informative, provided that the decision module is allowed to adapt. Such behavior is intuitive in the \textbf{\textit{A\_class}} scenario, where the input distribution itself remains unchanged, despite shifts in class composition. In the \textbf{\textit{A\_loc}} setting, although the temporal position of the discriminative pattern changes, the nature and scale of the signal remain consistent, and the drift range remains limited, enabling the decision module to adjust its policy without requiring an update of the underlying representation.

\begin{table}[!htb]
    \centering
    \caption{(\textit{a}): The mean percentage difference between the cumulative cost of full updated DQeND and the partially frozen version, over five random seeds. (\textit{b}): The mean of the median percentage difference of the hold-out \textit{AvgCost} drift of the full updated DQeND and partially frozen version, over five random seeds. Negative figures indicate degraded performance compared to original DQeND.}
    \begin{tabular}{l|cc|cc}
         & \multicolumn{2}{c|}{\textit{freeze} \textbf{encoder}} & \multicolumn{2}{c}{\textit{freeze} \textbf{decision}} \\
         & (\textit{a}) & (\textit{b}) & (\textit{a}) & (\textit{b}) \\
        \hline
        \textbf{\textit{I\_loc}} & $-32.42$\% & $-36.44$\% & $-36.61$\% & $-31.87$\% \\
        \textbf{\textit{I\_noise}} & $-11.29$\% & $-4.04$\% & $-19.28$\% & $-6.63$\% \\
        \hline
        \textbf{\textit{A\_loc}} & $1.38$\% & $-0.34$\% & $-43.52$\% & $-45.30$\% \\
        \textbf{\textit{A\_class}} & $1.32$\% & $2.14$\% & $-18.18$\% & $-4.39$\% \\
        \hline
    \end{tabular}
    \label{tab:ablation}
\end{table}

\section{Conclusion} \label{conclusion}

In this paper, we investigated the relevance of end-to-end learning for Early Classification of Time Series under evolving time series distributions. To address this question, we conducted extensive experiments within a controlled synthetic drift framework, systematically comparing end-to-end approaches against the widely used separable paradigm. To this end, we proposed DQeND, a Reinforcement Learning–based end-to-end ECTS architecture designed to remain closely comparable to state-of-the-art separable baselines. Across a broad range of gradual and abrupt non-stationarities, end-to-end approaches consistently demonstrated stronger robustness and adaptation capabilities than separable competitors, whose behavior often remained unstable and limited under drift. In particular, DQeND achieved strong online performance while preserving adapted behaviors on the original training concept, highlighting its ability to balance adaptation and retention in dynamic environments. More broadly, our results emphasize the benefits of jointly optimizing representation learning and triggering decisions in the context of ECTS.

Future work could include a more rigorous theoretical analysis of the stability–plasticity trade-off in ECTS, as well as the study of additional forms of non-stationarity, such as drifting cost functions or evolving class priors. Finally, beyond the controlled experimental setting considered in this work, the proposed framework provides a practical foundation for deployment in more complex real-world streaming environments.

\appendix

\section{Variability report}

In this section, the evaluation metrics displayed in Section \ref{sec_results} are detailed in Tables \ref{tab:variance1}, \ref{tab:variance2}, \ref{tab:variance3}  and variability is reported.

\begin{table}[!htb]
    \centering
    \caption{The mean cumulative cost and standard deviation, over five random seeds. \textbf{Bold} value indicates the best average performing method.}
    \resizebox{.48\textwidth}{!}{
    \begin{tabular}{l|cccc}
         & \textbf{\textit{I\_loc}} &  \textbf{\textit{I\_noise}} & \textbf{\textit{A\_loc}} & \textbf{\textit{A\_class}} \\
        \hline
         Alert(\textit{online}/\textit{freeze}) & $29.02 \pm 4.08$ & $46.14 \pm 0.3$ & $31.06 \pm 11$ & $32.15 \pm 1.27$\\
         Alert(\textit{online}/\textit{online}) & $31.87 \pm 5.49$ & $37.92 \pm 0.83$ & $23.32 \pm 1.88$ & $30.86 \pm 0.2$ \\
         DQeND & $\textbf{16.18} \pm 3.40$ & $\textbf{36.2} \pm 1.04$ & $\textbf{18.86} \pm 0.55$ & $32.98 \pm 3.88$\\
         EARLIEST & $20.53 \pm 6.54$ & $43 \pm 3.37$ & $19.90 \pm 0.79$ & $32.44 \pm 5.88$\\
         ELECTS & $17.33 \pm 1.81$ & $38.63 \pm 1.98$ & $29.98 \pm 12.06$ & $\textbf{25.79} \pm 3.52$\\
        \hline
    \end{tabular}}
    \label{tab:variance1}
\end{table}

\begin{table}[!htb]
    \centering
    \caption{The mean hold-out \textit{AvgCost} drift and standard deviation, over all the deployment steps and five random seeds. \textbf{Bold} value indicates the best average performing method.}
    \resizebox{.48\textwidth}{!}{
    \begin{tabular}{l|cccc}
         & \textbf{\textit{I\_loc}} &  \textbf{\textit{I\_noise}} & \textbf{\textit{A\_loc}} & \textbf{\textit{A\_class}} \\
        \hline
         Alert(\textit{online}/\textit{freeze}) & $0.3 \pm 0.04$ & $0.46 \pm 0.003$ & $0.31 \pm 0.11$ & $0.32 \pm 0.01$\\
         Alert(\textit{online}/\textit{online}) & $0.31 \pm 0.07$ & $0.38 \pm 0.009$ & $0.23 \pm 0.02$ & $0.31 \pm 0.002$ \\
         DQeND & $\textbf{0.16} \pm 0.03$ & $\textbf{0.36} \pm 0.01$ & $\textbf{0.19} \pm 0.01$ & $0.33 \pm 0.04$\\
         EARLIEST & $0.2 \pm 0.06$ & $0.43 \pm 0.03$ & $0.2 \pm 0.01$ & $0.32 \pm 0.06$\\
         ELECTS & $0.17 \pm 0.02$ & $0.39 \pm 0.02$ & $0.3 \pm 0.12$ & $\textbf{0.26} \pm 0.04$\\
        \hline
    \end{tabular}}
    \label{tab:variance2}
\end{table}

\begin{table}[!htb]
    \centering
    \caption{The mean hold-out \textit{AvgCost} original and standard deviation, over all the deployment steps and five random seeds. \textbf{Bold} value indicates the best average performing method.}
    \resizebox{.48\textwidth}{!}{
    \begin{tabular}{l|cccc}
         & \textbf{\textit{I\_loc}} &  \textbf{\textit{I\_noise}} & \textbf{\textit{A\_loc}} & \textbf{\textit{A\_class}} \\
        \hline
         Alert(\textit{online}/\textit{freeze}) & $0.31 \pm 0.02$ & $0.31 \pm 0.04$ & $0.21 \pm 0.04$ & $\textbf{0.37} \pm 0.03$\\
         Alert(\textit{online}/\textit{online}) & $0.33 \pm 0.06$ & $0.33 \pm 0.02$ & $0.18 \pm 0.02$ & $0.37 \pm 0.003$ \\
         DQeND & $\textbf{0.09} \pm 0.04$ & $\textbf{0.31} \pm 0.01$ & $\textbf{0.11} \pm 0.02$ & $0.4 \pm 0.04$\\
         EARLIEST & $0.27 \pm 0.07$ & $0.36 \pm 0.01$ & $0.11 \pm 0.07$ & $0.51 \pm 0.01$\\
         ELECTS & $0.14 \pm 0.07$ & $0.33 \pm 0.01$ & $0.19 \pm 0.1$ & $0.48 \pm 0.02$\\
        \hline
    \end{tabular}}
    \label{tab:variance3}
\end{table}

\section{Runtime analysis}


Computation times are reported in Table \ref{tab:runtime}. Overall, ELECTS and EARLIEST are the fastest approaches, both during training and deployment. Unlike Alert and DQeND, which rely on convolution-based encoders, these methods are LSTM-based and therefore benefit from the CPU-only experimental setup used in this work. In practice, convolutional approaches would likely reduce this gap when leveraging GPU acceleration. For Alert-based methods indeed, most of the training time is spent fitting the ELM-based classifier rather than the trigger module itself, confirming the relatively low computational overhead of the RL trigger. In contrast, DQeND exhibits the highest training and update times.

More broadly, the deployment times of DQeND remain significantly longer. This is consistent with the known sample inefficiency of value-based RL algorithms, whose repeated updates and replay-based optimization can constitute computational limitations in online settings.

\begin{table}[!htb]
    \centering
    \caption{Computation times (in \textit{s.}). The deployment are displayed on a \textit{per batch} basis. The total time accounts for the all 100 deployment steps. The fit time of the Alert-based methods are decomposed over (\textit{classifier} fit time, \textit{trigger} fit time).}
    \resizebox{0.48\textwidth}{!}{
    \begin{tabular}{l|c|cc|c}
         & \textbf{Training} & \multicolumn{2}{c|}{\textbf{Deployment} (\textit{per batch})} & \textbf{Total} $\downarrow$ \\
         & Fit & Inference & Update &  \\
         \hline
         ELECTS & 31.89 & 0.026 & 0.05 & 39.22 \\
         EARLIEST & 119.25 & 0.23 & 0.09 & 138.98 \\
         Alert (\textit{online}/\textit{online}) & (78.02, 46.55) & 2.39 & 2.45 & 612.75 \\
         Alert (\textit{retrain}/\textit{retrain}) & (78.02, 46.55) & 2.71 & 3.59 & 759.86 \\
         DQeND & 248.74 & 2.90 & 5.75 & 1124.31 \\
        \hline
    \end{tabular}}
    \label{tab:runtime}
\end{table}


\bibliographystyle{ACM-Reference-Format}
\bibliography{biblio}

 \end{document}